# Simultaneous Location of Rail Vehicles and Mapping of Environment with Multiple LiDARs

Yusheng Wang, Weiwei Song, Yidong Lou, Fei Huang, Zhiyong Tu and Shimin Zhang

*Abstract*—Precise and real-time rail vehicle localization as well as railway environment monitoring is crucial for railroad safety. In this letter, we propose a multi-LiDAR based simultaneous localization and mapping (SLAM) system for railway applications. Our approach starts with measurements preprocessing to denoise and synchronize multiple LiDAR inputs. Different frame-to-frame registration methods are used according to the LiDAR placement. In addition, we leverage the plane constraints from extracted rail tracks to improve the system accuracy. The local map is further aligned with global map utilizing absolute position measurements. Considering the unavoidable metal abrasion and screw loosening, online extrinsic refinement is awakened for long-during operation. The proposed method is extensively verified on datasets gathered over 3000 km. The results demonstrate that the proposed system achieves accurate and robust localization together with effective mapping for large-scale environments. Our system has already been applied to a freight traffic railroad for monitoring tasks.

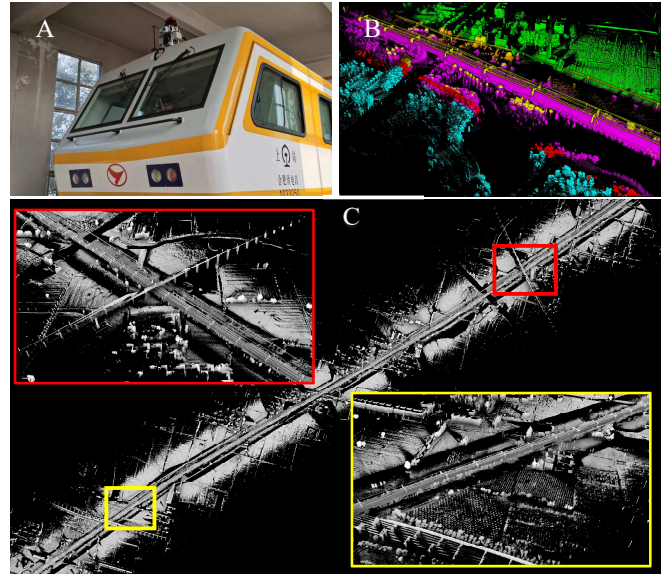

Fig. 1. A): The hardware setup of seven LiDARs on the rail vehicle (5 Livox Horizon, 81.7° × 25.1° FoV, 2 Livox Tele-15, 14.5° × 16.2° FoV). B): The mapping result of the proposed system coded by LiDAR IDs. C): An example of the real-time mapping result from our proposed framework, the color is coded by PCV. Both the map of railroad and the environment (up to 500 m from the rail vehicle) can be well acquired with the cooperation of seven LiDARs.

## I. INTRODUCTION

MULTI-MODAL sensor integration has become a crucial prerequisite for the real-world navigation systems. Recent studies have reported successful deployment of LiDAR-inertial system in handled devices [1], legged robots [2], unmanned ground and aerial vehicles (UAV and UGV) [3], [4], autonomous cars [5], and boats [6]. However, the LiDAR-inertial based simultaneous localization and mapping (SLAM) is still an open problem for rail vehicle applications.

With the framework of estimating train state and mapping the surrounding in the meantime, SLAM is a promising approach towards rail vehicle localization and railroad environment monitoring problems. However, a number of difficulties affect the application of SLAM on rail vehicles:

*Large velocities*: Existing SLAM frameworks are usually evaluated with slow motion platforms. For example, the maximum speed of UAV [4] and the Clearpath Jackal [7] is below 10 km/h. On the contrary, the average speed of the slowest rail vehicle is already beyond 80 km/h. The great speed not only cause motion blur but also introduce extra challenge for frame-to-frame registration of small FoV LiDARs [8], [9].

*No-revisited districts*: The accumulated drift of most SLAM algorithms can be corrected with detected loops at backend. However, there are no revisited districts for the railroad.

*Long-during repetitive scenes*: The safety regulations require a clearance gauge for the railroad, where only the rail tracks, and track side infrastructures are observable, making most of the railroad with repetitive structures and prone to degenerate.

*Large mapping coverage required*: In practice, many railroad contingencies are the consequence of insufficient environment awareness and prevention. For example, some power failures are caused by short circuit from blown away greenhouse plastic film. Therefore, a good knowledge of both the track bed and the long-range environment is required, leading to a multi-LiDAR setup on rail vehicles, result in extra consideration of precise extrinsic calibration problems.

To tackle these challenges, in this letter, we present a multi-LiDAR based localization and mapping system for rail vehicles, with the system setup and some real-time mapping result shown in Fig. 1. This system receives measurements from multiple LiDARs, an IMU, rail vehicle odometer, and a global navigation satellite system (GNSS) receiver. All the point clouds are first denoised and made distortion-free utilizing IMU/odometer information. According to the various installment and FoV of LiDARs, different LiDAR odometry methods are employed. In addition, we leverage the typical geometric patterns on railroads to further refine the pose estimation. The local maps are then registered to global with absolute positioning data and compass heading. In summary, our contributions are:

1) We propose a framework that tightly fuses LiDAR, IMU, rail vehicle wheel odometer, and GNSS through sliding window based factor graph formulation.
2) We employ different scan matching methods according to the geometric displacement of different LiDARs. Besides, we introduce an online extrinsic estimating and updating scheme for long-during tasks.
3) We leverage the geometric structure of environment for state estimation, where plane constraints from extracted rail tracks and height information descriptor are employed to prevent degeneracy.

## II. Related Work

### A. Train Positioning and Railway Mapping Solutions

The existing train positioning strategy is mainly dependent on trackside infrastructures like track circuits [10] and Balises [11]. Since the accuracy of these systems is determined by the operation interval, they are neither accurate nor efficient for intelligent rail transportation systems. Considering its large capital investment and low efficiency, many researchers seek to complement the system limitations with onboard sensors. The satellite-based methods utilizes the GNSS for train positioning, and the accuracy can be further improved with integrated track odometry [12], wheel odometer [13], and IMU [14]. However, these methods merely achieve train state information without awareness of environmental information.

The current railroad environment monitoring is still a human-intensive work, and a professional technician need to go with the train driver every time to manually check the infrastructure defects. Although the visual approaches [15], [16] have been largely investigated, they are inaccurate for range measuring and sensitive to illumination conditions. In many of the previous works [17]–[19], laser scanners have been included in the mobile mapping system (MMS) for railroad monitoring tasks. As a direct geo-referencing approach, the MMS system requires high-precision GNSS/IMU determination and survey-grade laser scanners. Although these solutions can achieve highly-accurate 3D maps, they are costly for large deployment and less-efficiency for real-time perception.

The potential of SLAM for rail vehicles localization and mapping has not been well investigated. One of the early works, RailSLAM, jointly estimated the train state and validated the correctness of initial track map based on a general Bayesian theory [20]. The performance of Visual-inertial odometry on rail vehicles have been extensively evaluated in [21], indicating that the Visual-inertial odometry is not reliable for railroad applications. But the LiDAR-inertial SLAM is still an open problem for railway applications.

### B. Multi-LiDAR Based SLAM

According to the data association scheme, multiple LiDAR integration can be classified into centralized and decentralized approaches. A centralized multi LiDAR method is presented in [5], this approach runs onboard with several desirable features, including a tightly-coupled multi-LiDAR motion estimation, online extrinsic calibration with convergence identification, and uncertainty-aware mapping. However, as a LiDAR-only SLAM, this approach is inevitable to long-duration navigation drifts. A decentralized framework based on the Extended Kalman filter (EKF) is proposed in [22]. This method distribute the intensive computation among dedicated LiDARs, and treats each LiDAR input as independent modules for pose estimation. Although approval accuracy can be reached, this system is only simulated on a high-performance computer, the communication delay and message loss in the real case are not taken into consideration.

Six LiDARs are integrated into state estimation and map construction in our previous work [23], achieving accurate result in complex urban roads. And this paper seeks to achieve real-time, low-drift and robust odometry and mapping for large-scale railroad environments with multiple LiDAR integrated LiDAR-inertial SLAM.

## III. System Overview

As shown in Fig. 2, the proposed system receives measurements from seven LiDARs, an IMU, wheel odometers, a GNSS receiver and outputs 10 Hz odometry as well as 1Hz mapping. In addition, the multiple LiDAR placement is illustrated in Fig. 3. before dive into the details of the methodology, we first define the notations used throughout this article.

We denote $(\cdot)^W$, $(\cdot)^B$, $(\cdot)^L$, and $(\cdot)^O$ as the world, body, LiDAR and odometer frame. In addition, we define $(\cdot)_W^B$ as the transform from world frame to the IMU frame. We use both rotation matrix $\mathbf{R}$ and quaternion $\mathbf{q}$ to represent rotation. Besides, we denote $\otimes$ as the multiplication between two quaternions. $(\hat{\cdot})$ is denoted as the estimation of a certain quantity.

### A. State Definition

We split the full state vector $\boldsymbol{\chi}$ into three groups, with:

$$\boldsymbol{\chi} = [\boldsymbol{\chi}_s, \quad \boldsymbol{\chi}_v, \quad \boldsymbol{\chi}_e]$$
$$= [\mathbf{x}_1, \dots, \mathbf{x}_l, \mathbf{x}_{l+1}, \dots, \mathbf{x}_{N+1}, \mathbf{x}_2^p, \dots, \mathbf{x}_{l_7}^p] \quad (1)$$
$$\mathbf{x}_i = [\mathbf{p}_i, \mathbf{v}_i, \mathbf{q}_i, \mathbf{b}_a, \mathbf{b}_g, \mathbf{c}_i], i \in [1, N+1]$$
$$\mathbf{x}_{l_i}^c = [\mathbf{p}_{l_i}^p, \mathbf{q}_{l_i}^p], i \in [2,6]$$

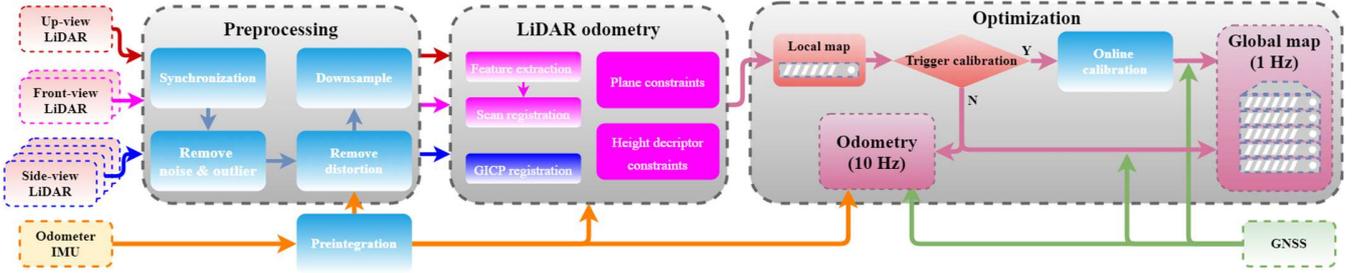

**Fig. 2.** Block diagram illustrating the full pipeline. The seven input LiDARs are synchronized, denoised, made distortion free and downsampled at the preprocessing. The two front-view LiDARs (*1* and *2*) perform feature extraction and scan registration for LiDAR odometry, besides, the rail track plane constraints as well as height descriptor constraints are extracted. For the side-view LiDARs (*3 ~ 6*) with small FoV, they only employ generalized iterative closest points (GICP) for registration. The online calibration is triggered when the vehicle travels for a relative long time. All the constraints are jointly optimized with constructed factor graph. For the up-view LiDAR (*7*), it only employs the simultaneous odometry output for mapping and calibration.

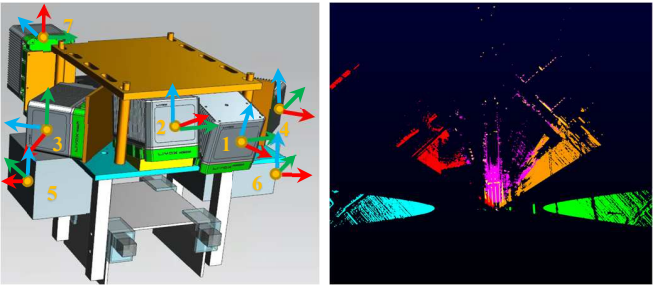

**Fig. 3.** Visual illustration of the placement of seven LiDARs (left), including 2 front-view Livox Horizon (*1*, *2*), 2 side-view Livox Horizon (*3*, *4*), 2 side-view Livox Tele-15 (*5*, *6*), and 1 up-view Livox Horizon (*7*). The red-green-blue color indicates relative *x-y-z* axis. A single scan coded by LiDAR IDs (right).

where $\chi_s = [\mathbf{x}_1, ..., \mathbf{x}_l]$ are considered as the solid states, with accurate extrinsic. $\mathbf{x}_i$ is the state of the primary LiDAR (LiDAR *1*), with $\mathbf{p} \in \mathbb{R}^3$, $\mathbf{v} \in \mathbb{R}^3$, and $\mathbf{q} \in SO(3)$ denoting the position, linear velocity, and orientation vector. $\mathbf{b}_a$ and $\mathbf{b}_g$ are the usual IMU gyroscope and accelerometer biases. And **c** is the scale factor of the odometer. As the rail vehicles work for long hours, the unavoidable abrasion and deformation will shift the original parameters. We hereby introduce the states with variations, $\chi_v = [\mathbf{x}_{l+1}, ..., \mathbf{x}_{N+1}]$ and the dedicated LiDAR extrinsic $\chi_e = [\mathbf{x}_{l_2}^p, ..., \mathbf{x}_{l_7}^p]$ for online refinement, and $\mathbf{x}_{l_i}^p$ represents the extrinsics from the auxiliary LiDAR *i* to the primary LiDAR.

*B. Maximum-a-Posterior Problem*

We seek to estimate the trajectory and map the surrounding of a rail vehicle with multi-sensor measurements, in which the state estimation procedure can be formulated as a maximum-a-posterior (MAP) problem. Given the measurements $\mathcal{Z}_k$ and the history of states $\chi_k$, the MAP problem can be formulated as:

$$\chi_k^* = \underset{\chi_k}{\operatorname{argmax}}\, p(\chi_k|\mathcal{Z}_k) \propto p(\chi_0)p((\mathcal{Z}_k|\chi_k)) \quad (2)$$

If the measurements are conditionally independent, then (2) can be solved through least squares minimization:

$$\chi^* = \underset{\chi_k}{\operatorname{argmin}} \sum \sum_{i=1}^{k} \|r_i\|^2 \quad (3)$$

where $r_i$ is the residual of the error between the predicted and measured value.

*C. Optimization*

Since the extrinsics are configured with a total station and a 3D laser scanner before each experiment, we assume the LiDAR extrinsics are accurate enough at the beginning, and the optimal solid states $\chi_s$ are obtained through minimizing:

$$\mathcal{F}_{\mathcal{M}}(\chi) = \mathcal{F}_{\mathcal{M}}(\chi_s)$$
$$= \min_{\chi_s}\{\|r_p\|^2 + \sum_{i=1}^{N_K}\|r_{\mathcal{I}_i}\|^2 + \sum_{i=1}^{N_{\mathcal{L}_K}} r_{\mathcal{L}_i}$$
$$+ \sum_{i=1}^{N_{\mathcal{R}_K}} r_{\mathcal{R}_i} + \sum_{i=1}^{N_{\mathcal{P}_K}}\|r_{\mathcal{P}_i}\|^2 + \sum_{i=1}^{N_{\mathcal{G}_K}}\|r_{\mathcal{G}_i}\|^2\} \quad (4)$$

where $r_p$ is the prior factor marginalized by Schur-complement [24], $r_{\mathcal{I}_i}$ is the residual of IMU/odometer preintegration result. $r_{\mathcal{L}_i}$, $r_{\mathcal{R}_i}$ and $r_{\mathcal{P}_i}$ defines the residual of feature-based and GICP-based scan registration, as well as the residual of ground constraints. The residual of global positioning system is $r_{\mathcal{G}_i}$.

Once the rail vehicle runs for a long time, the online extrinsic calibration is triggered. And we exploit the map registration based measurements to correct the extrinsics:

$$\mathcal{F}_{\mathcal{M}}(\chi) = \mathcal{F}_{\mathcal{M}}(\chi_v) + \mathcal{F}_{\mathcal{M}}(\chi_e) \quad (5)$$

IV. METHODOLOGY

*A. Calibration*

Considering the highly restricted FoV and non-overlapping of our multi-LiDAR system setup, we utilize a total station and a 3D laser scanner to achieve the preliminary LiDAR extrinsics through EPnP [25].

Unlike many data gathering vehicles which only works for one or two hours each time, the minimum operation time for a maintenance rail vehicle is five hours from our experience, with the longest one continuously runs for three days, covering thousands of kilometers. Since the metal abrasion and seasonal deformation is unavoidable for long-during tasks, we refine the extrinsics for two criterias.

*Stop at certain stations*: The rail vehicles need to stop at some certain stations to wait for the dispatching orders from the automatic control system (ATC). Since the railway stations are with many column-like pillars and man-made structures, we employ the edge-based camera-LiDAR calibration algorithm [26] to refine the parameters between multiple LiDARs and the panoramic camera.

*Long during*: Once the rail vehicle runs for a long time without stop, the online extrinsic calibration is triggered, and we leverage (5) for refinement.

### B. IMU/Odometer Preintegration Factor

The raw accelerometer, gyroscope, and train wheel odometer measurements, $\hat{\mathbf{a}}$, $\hat{\boldsymbol{\omega}}$, and $\hat{\mathbf{v}}^O$ are given by:

$$\hat{\mathbf{a}}_k = \mathbf{a}_k + \mathbf{R}_W^{B_k}\boldsymbol{g}^W + \mathbf{b}_{a_k} + \boldsymbol{\eta}_a$$
$$\hat{\boldsymbol{\omega}}_k = \boldsymbol{\omega}_k + \mathbf{b}_{\omega_k} + \boldsymbol{\eta}_\omega$$
$$c^{O_k}\hat{\mathbf{v}}^O = \mathbf{v}^O + \boldsymbol{\eta}_o \qquad (6)$$

where $\boldsymbol{g}^W = [0,0,g]^T$ is the gravity vector in the world frame. $\boldsymbol{\eta}_a$, $\boldsymbol{\eta}_\omega$, and $\boldsymbol{\eta}_o$ are the zero-mean white Gaussian noise.

Given two consecutive frames $k$ and $k+1$, the position, velocity, and orientation states can be propagated by the IMU/odometer measurements with:

$$\mathbf{p}_{B_{k+1}}^W = \mathbf{p}_{B_k}^W + \mathbf{v}_{B_k}^W \Delta t_k + \iint_{t=t_k}^{t_{k+1}} (\mathbf{R}_t^W(\hat{\mathbf{a}}_t - \mathbf{b}_{a_t} - \boldsymbol{\eta}_a) - \boldsymbol{g}^W) dt^2$$

$$\mathbf{v}_{B_{k+1}}^W = \mathbf{v}_{B_k}^W + \int_{t=t_k}^{t_{k+1}} (\mathbf{R}_t^W(\hat{\mathbf{a}}_t - \mathbf{b}_{a_t} - \boldsymbol{\eta}_a) - \boldsymbol{g}^W) dt$$

$$\mathbf{q}_{B_{k+1}}^W = \mathbf{q}_{B_k}^W \otimes \int_{t=t_k}^{t_{k+1}} \frac{1}{2}\Omega(\hat{\boldsymbol{\omega}}_t - \mathbf{b}_{\omega_t} - \boldsymbol{\eta}_\omega)\mathbf{q}_t^{B_k} dt \qquad (7)$$

where

$$\Omega(\boldsymbol{\omega}) = \begin{bmatrix} -[\boldsymbol{\omega}]_\times & \boldsymbol{\omega} \\ -\boldsymbol{\omega}^T & 0 \end{bmatrix}, [\boldsymbol{\omega}]_\times = \begin{bmatrix} 0 & -\omega_z & \omega_y \\ \omega_z & 0 & -\omega_x \\ -\omega_y & \omega_x & 0 \end{bmatrix} \qquad (8)$$

Based thereupon and the preintegration form in [24], we can formulate the IMU and odometer increment between $k$ and $k+1$ as:

$$\boldsymbol{\alpha}_{B_{k+1}}^{B_k} = \iint_{t=k}^{k+1} \mathbf{R}_{B_t}^{B_k}(\hat{\mathbf{a}}_t - \mathbf{b}_{a_t} - \boldsymbol{\eta}_a) dt^2$$

$$\boldsymbol{\beta}_{B_{k+1}}^{B_k} = \int_{t=k}^{k+1} \mathbf{R}_{B_t}^{B_k}(\hat{\mathbf{a}}_t - \mathbf{b}_{a_t} - \boldsymbol{\eta}_a) dt$$

$$\boldsymbol{\gamma}_{B_{k+1}}^{B_k} = \int_{t=k}^{k+1} \frac{1}{2}\Omega(\hat{\boldsymbol{\omega}}_t - \mathbf{b}_{\omega_t} - \boldsymbol{\eta}_\omega)\boldsymbol{\gamma}_{B_t}^{B_k} dt$$

$$\boldsymbol{\alpha}_{O_{k+1}}^{O_k} = \int_{t=k}^{k+1} \mathbf{R}_{O_t}^{O_k}(c^{O_k}\hat{\mathbf{v}}^O - \boldsymbol{\eta}_s o) dt \qquad (9)$$

Using the calibration parameter $\mathbf{p}_{O_{k+1}}^{B_{k+1}}$ between the odometer and the IMU measured by a total station, we can also transform $\boldsymbol{\alpha}_{O_{k+1}}^{O_k}$ into IMU frame $\boldsymbol{\phi}_{B_{k+1}}^{B_k}$ with:

$$\boldsymbol{\phi}_{B_{k+1}}^{B_k} = \int_{t=k}^{k+1} \mathbf{R}_{B_t}^{B_k}\mathbf{R}_{O_t}^{B_t}(c^{O_k}\hat{\mathbf{v}}^O - \boldsymbol{\eta}_s o) dt \qquad (10)$$

Finally, the residual of preintegrated IMU/odometer data $\begin{bmatrix}\delta\boldsymbol{\alpha}_{B_{k+1}}^{B_k} & \delta\boldsymbol{\beta}_{B_{k+1}}^{B_k} & \delta\boldsymbol{\theta}_{B_{k+1}}^{B_k} & \delta\mathbf{b}_a & \delta\mathbf{b}_g & \delta\boldsymbol{\phi}_{B_{k+1}}^{B_k} & \delta c^O\end{bmatrix}^T$ is given as:

$$r_J(\hat{\mathcal{Z}}_{B_{k+1}}^{B_k},\chi) = \begin{bmatrix} \mathbf{R}_W^{B_k}\left(\mathbf{p}_{B_{k+1}}^W - \mathbf{p}_{B_k}^W + \frac{1}{2}\boldsymbol{g}^W \Delta t_k^2 - \mathbf{v}_{B_k}^W \Delta t_k\right) - \hat{\boldsymbol{\alpha}}_{B_{k+1}}^{B_k} \\ \mathbf{R}_W^{B_k}\left(\mathbf{v}_{B_{k+1}}^W + \boldsymbol{g}^W \Delta t_k - \mathbf{v}_{B_k}^W\right) - \hat{\boldsymbol{\beta}}_{B_{k+1}}^{B_k} \\ 2\left[\left(\mathbf{q}_{B_k}^W\right)^{-1} \otimes \left(\mathbf{q}_{B_{k+1}}^W\right) \otimes \left(\hat{\boldsymbol{\gamma}}_{B_{k+1}}^{B_k}\right)^{-1}\right]_{2:4} \\ \mathbf{b}_{a_{k+1}} - \mathbf{b}_{a_k} \\ \mathbf{b}_{g_{k+1}} - \mathbf{b}_{g_k} \\ \mathbf{R}_W^{B_k}\left(\mathbf{p}_{B_{k+1}}^W - \mathbf{p}_{B_k}^W + \mathbf{R}_{B_{k+1}}^W \mathbf{p}_{O_{k+1}}^{B_{k+1}}\right) - \hat{\boldsymbol{\phi}}_{B_{k+1}}^{B_k} \\ c^{O_{k+1}} - c^{O_k} \end{bmatrix} \qquad (11)$$

where $[\cdot]_{2:4}$ is used to take out the last four elements from a quaternion.

### C. LiDAR Odometry Factors

Since the range measuring error in the axial direction is large for short-distance, we first remove the too close points from LiDAR. Then we apply the IMU/odometer increment model to correct LiDAR point motion distortion with linear interpolation.

As shown in Fig. 3, we select the LiDAR *1* as the primary unit $\mathcal{L}^1$, and others are regarded as secondary LiDARs $\mathcal{L}^i, i \in [2,7]$. The scan period of the primary LiDAR is approximately 0.1 s, thus for any $k$, the time-span between $t_k$ and $t_{k+1}$ is also 0.1 s. However, the secondary LiDARs are less likely to have the identical timestamps with the primary unit due to unpredictable time delays and information loss. As they are sent to different threads for feature extraction, we then merge all the feature points whose start time fall in $[t_k, t_{k+1})$ to obtain fused scan $\mathcal{F}_k$. The combined points inherit the timestamp of the primary LiDAR and stretch over the time span $[t_k, t'_k)$. The IMU measurements in the interval $[t_{k-1}, t'_k]$ are used for state propagation, where the samples in $[t_{k-1}, t_k)$ are utilized for preintegration and the other subsets are utilized for motion compensation.

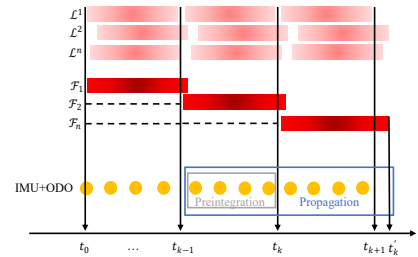

**Fig. 4.** Illustration of the synchronization and feature extraction within multiple LiDAR scans

For LiDAR *1* and *2*, we follow the work of [27] to extract two sets of feature points from denoised and distortion-free point cloud. The edge features $\varepsilon$ are selected with high curvature and the planar features $\rho$ are with low curvature. Then we take the fused feature points to perform scan registration with the edge and planar patch correspondence computed through point-to-line and point-to-plane distances, $d_{\varepsilon 2\varepsilon}$ and $d_{\rho 2\rho}$. Then the LiDAR odometry residual at k-th frame can be formulated by:

$$r_{\mathcal{L}_k} = \sum_{i=1}^{N_\mathcal{L}} w_i r_{\mathcal{L}_i}$$

$$r_{\mathcal{L}_i} = \sum_{j=1}^{N_\varepsilon}(d_{\varepsilon 2\varepsilon_j})^2 + \sum_{j=1}^{N_\rho}(d_{\varepsilon 2\varepsilon_j})^2 \quad (12)$$

where $N_\mathcal{L}$ denotes the number of LiDARs. $w_i$ is the weighting factor for multiple LiDAR measurements evaluation, and can be expressed as:

$$w_i = w_i^I w_i^D$$
$$w_i^I = 1 - (\frac{\|\mathbf{p}_{B_k}^{B_{k+1}}\| - \|\mathbf{p}_{L_k}^{L_{k+1}}\|}{\|\mathbf{p}_{B_k}^{B_{k+1}}\|})^2$$
$$w_i^D = \frac{\lambda_i}{\lambda_{emp}} \quad (13)$$

where $w_i^I$ is the inertial weighting factor, $\mathbf{p}_{B_k}^{B_{k+1}}$, $\mathbf{p}_{L_k}^{L_{k+1}}$ are the IMU/odometer preintegration and the LiDAR odometry pose divergence between two consecutive keyframes. $w_i^D$ denotes the degeneracy-aware weight factor, with the degeneracy factor calculated following [28]. The empirical threshold $\lambda_{emp}$ is get from feature-rich railway stations.

Since the horizontal FoV for the four side-view LiDARs are poorly restricted (25.1° for the Horizon, 14.5° for the Tele-15), the feature-based scan matching is prone to fail for large speed rail vehicles. We hereby leverage the GICP-based factor graph optimization [29] to get $r_{\mathcal{R}_k}$.

We notice that the LiDAR-only odometry with LiDAR of limited FoV is over-sensitive to the vibrations caused by the joint of rail tracks and the rail track turnouts, where errors may appear in the pitch direction. Besides, the two rail tracks are not of the same height at turnings, and the LiDAR-only odometry will keep this roll divergence even in the following straight railways. Illustrated in [30], the planar features from segmented ground can effectually constrain the roll and pitch rotation. However, the angle-based ground extraction is not robust for railways as the small height variations will be ignored by the segmentation, which will generate large vertical divergence for large-scale mapping tasks.

We hereby employ the rail track plane to provide ground constraints. We first detect the track bed area using the LiDAR

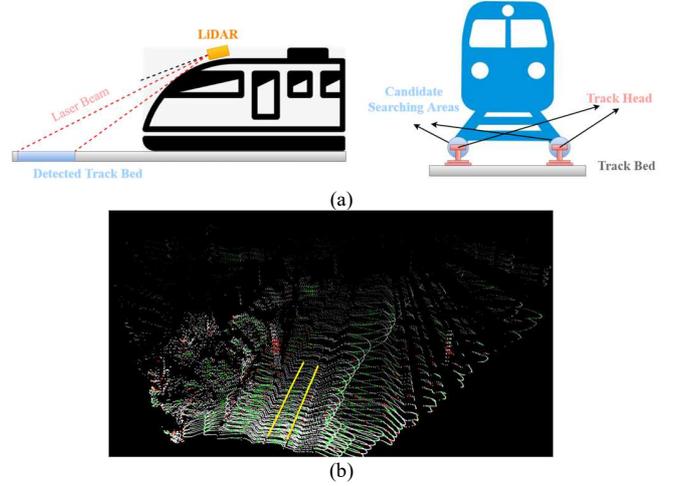

**Fig. 5.** (a) Illustration of the track bed area detection and candidate rail track points searching. (b) The extracted planar points (green), edge points (red), and rail tracks (yellow) from the LiDAR *1*.

sensor mounting height and angle as illustrated in Fig. 5 (a). With the assumption of the LiDAR is centered between two rail tracks, we can set two candidate areas around the left and right rail tracks and search the points with local maximum height over the track bed. Two straight lines can then be fixed using random sample consensus (RANSAC) [31] method. Finally, we exploit the idea of region growing [32] for further refinement, with the result shown in Fig. 5 (b).

We are now able to define a plane with the two sets of rail track points using RANSAC. And the ground plane $m$ can be parameterized by the normal direction vector $n_p$ and a distance scalar $d_p$, $m = [n_p^T, d_p]^T$. Then the ground plane measurement residual can be expressed as:

$$r_{\mathcal{P}_k} = m_{k+1} - \mathbf{T}_{L_{k+1}}^{L_k} m_k \quad (14)$$

### D. Optimization with Online Calibration

Inspired by [5] and [22], we treat the online calibration as a submap registration problem. When the long-during operation (normally 4 hours) triggers the online calibration, the system will collect the submaps constructed under respective LiDAR coordinates, and perform the ICP algorithm to align the various coordinate frames. When the online calibration of LiDAR *1 ~ 6* is finished, LiDAR *7* will utilize the optimized pose for submap construction and follow-up calibration.

### E. GNSS Factor

The accumulated drifts of the system can be corrected using GNSS measurements. The GNSS factor is added when the estimated position covariance is larger than the reported GNSS covariance in [6]. However, we find that the reported GNSS covariance is not trustworthy sometimes, and may yield blurred or inconsequent mapping result. We hereby model the GNSS measurements $p_k^W$ with additive noise, and the global position residual can be defined as:

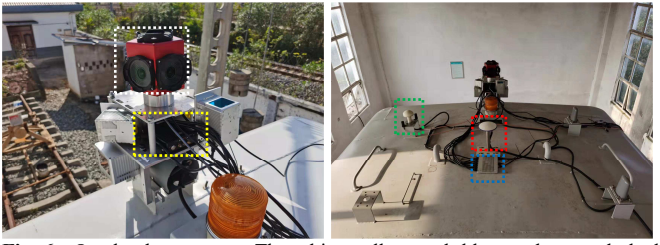

**Fig. 6.** Our hardware setup. The white, yellow, red, blue, and green dashed rectangle indicates the Ladybug5+, MiniII-D-INS and M39, localization antenna, Livox Hub, and GPS-PPS synchronization units, respectively.

$$r_{\mathcal{G}_k} = \mathbf{R}_W^{B_k}(\mathbf{p}^{W_k} - \mathbf{p}_W^B - \mathbf{p}_{B_k}^W$$
$$+ \frac{1}{2}\boldsymbol{g}^W \Delta t_k^2 - \mathbf{v}_{B_k}^W \Delta t_k) - \widehat{\boldsymbol{\alpha}}_{B_{k+1}}^{B_k} \quad (15)$$

where $\mathbf{p}_W^B$ is the transformation from the receiver antenna to the IMU, which can be obtained from installation configuration. Note that we only consider the single point positioning (SPP) result as input due to the inconsistent 4G communication quality for long railroads.

*E. Map Management*

The accurate scan-to-map registration of LOAM relies on the convergence of nonlinear optimization from sufficiently many iterations. However, we find the scan-to-map sometimes does not converge due to insufficient correspondences caused by large velocity, and destroy the whole mapping result. To cope with this problem, we propose a submap-based two-stage map-to-map registration, which first creates submaps based on local optimization, and utilizes the GNSS measurements for error correction and map registration. Once the number of iterations reaches a threshold, we introduce the GNSS positions as initial guess for ICP registration between current frame and the current accumulated submap. In addition, we also leverage the GNSS information for submap-to-submap registration using the normal distribution transform (NDT) [33]. In practice, 10 keyframes are maintained in each submap, which can reduce the mapping blurry caused by frequent correction.

## V. EXPERIMENT

The setup of the seven LiDARs is shown in Fig. 1 and Fig. 3, and the overall system is shown in Fig. 6. All the LiDARs are connected via a Livox Hub [1]. Besides, we employ Ladybug5+ [2] panoramic camera for calibration and LiDAR-camera based object detection. Additionally, the system also fuses an integrated navigation unit Femotomes MiniII-D-INS[3] and rail vehicle wheel odometers. All the sensors are hardware-synchronized with a u-blox EVK-M8T GNSS timing evaluation kit using GPS pulse per second (GPS-PPS).

We employ a personalized onboard computer, with i9-10980HK CPU (2.4 GHz, octa-core), 64GB RAM, for real-time processing. All our algorithms are implemented in C++ and executed in Ubuntu Linux using the ROS [34].

We conduct a series of experiments on different railroads, and we employ two datasets for explanation here as listed in TABLE I. The ground truth is kept by the post processing result of a MPSTNAV M39 GNSS/INS integrated navigation system[4] (with RTK corrections sent from Qianxun SI).

TABLE I.
DETAILS OF THE TWO SELECTED DATASETS

| Name | Duration (sec) | Distance (km) |
|---|---|---|
| FY-Back1 | 427 | 6.7 |
| HQ-to2 | 1643 | 27.3 |

*A. Result of Online Extrinsic Calibration*

The maintenance rail vehicle leaves the station at 6:57 AM and works until 3:32 PM, with *FY-Back1* covering a portion of the return data. Shown in Fig. 7 A, the mapping result is blurred without extrinsic refinement. After an online extrinsic calibration around 43 s, both the rotation and translation converge to a stable value. The clear and well-matched power towers (joint mapping of LiDAR *1*, *2*, *4*, and *6*) in Fig. 7 B show the remarkable ability of our algorithm to refine the extrinsic.

*B. Result of State Estimation*

We now present a series of evaluation to quantitatively analyze our proposed framework. We employ two novel Livox LiDAR based graph SLAM, Lili-om [3] and Lio-Livox[5] for comparison. Both of them directly take the calibrated and merged point clouds as input. Besides, the odometer and global constraints are also manually added for a fair comparison. TABLE II summarizes the root mean square error (RMSE) metrics. It is seen that Lio-Livox has a worse performance than SPP due to wrongly detected plane constraints, which generates large deviation in vertical direction. Besides, both algorithms cannot achieve real-time performance on a i9-11900K, 128GB RAM desktop.

TABLE II.
RMSE METRICS OF VARIOUS APPROACHES

| | Lili-om | Lio-Livox | SPP | Ours |
|---|---|---|---|---|
| FY-Back1 | 1.64 | 3.57 | 3.38 | **0.73** |
| HQ-to2 | 2.08 | 4.82 | 3.45 | **0.92** |

*C. Result of Multi-LiDAR Mapping*

We show that our proposed method is accurate enough to build large-scale map of railroad environments. The real-time mapping is shown in Fig. 8 and Fig. 9. We can see that the point cloud data from different LiDARs is aligned well together and the consistency is kept locally. The well-matched result with the satellite image indicates that our proposed method is of high precision globally. In addition, we leverage the refined camera-LiDAR extrinsic to plot the colored mapping result in Fig. 10.

---

[1] https://www.livoxtech.com/hub
[2] https://www.flir.com/products/ladybug5plus/
[3] http://www.femtomes.com/en/MiniII.php?name=MiniII
[4] http://www.whmpst.com/en/imgproduct.php?aid=29
[5] https://github.com/Livox-SDK/LIO-Livox

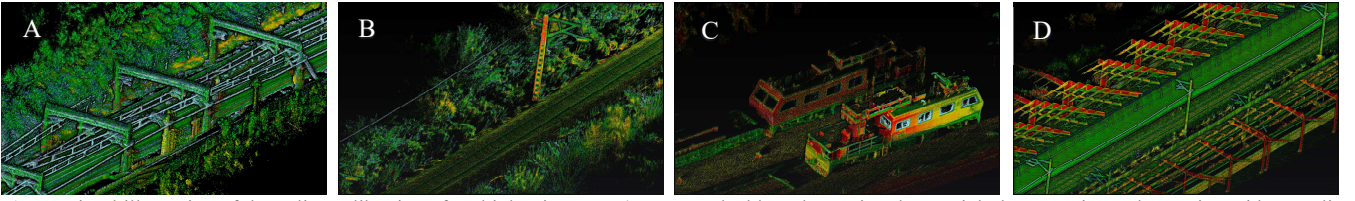

**Fig. 7.** Visual illustration of the online calibration of multiple LiDARs. A) presents the blurred mapping due to eight-hour continuously running without online refinement, where the power towers are 'stretched' and the up-down half does not coincide with each other. B) ~ D) presents the mapping of a power tower, two maintenance rail vehicles, and a station after online extrinsic refinement. All the color is coded by intensity variations.

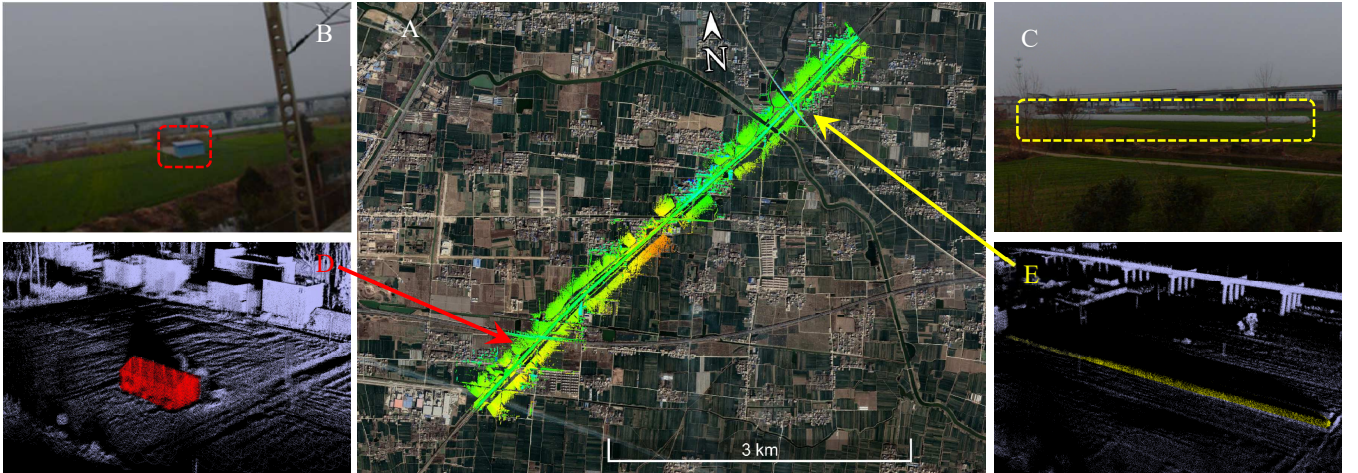

**Fig. 8.** A): The mapping result of *FY-Back1* aligned with the satellite map, and the color is coded by height variations. B) and C) indicates two examples of potential risks detected by the panoramic camera, with B) a prefabricated houses 110 m away from the railroad central line and C) a greenhouse plastic film 162 m away from the railroad central line. D) and E) denotes the corresponding point clouds of B) and C).

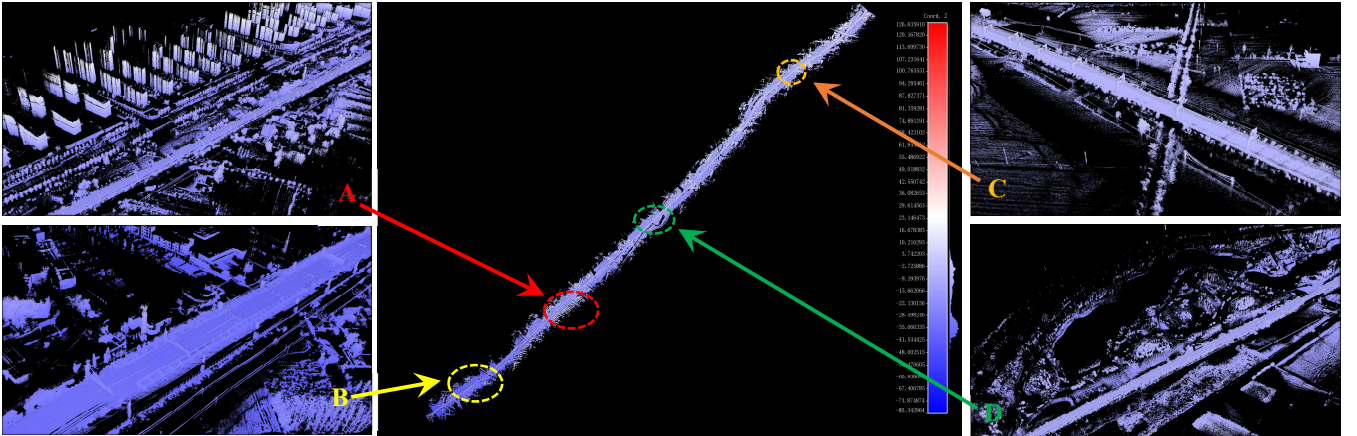

**Fig. 9.** The mapping result of *HQ-to2* coded by height variations. Note that the outlier points (with large *z* value) are caused by the direct sunlight. A) ~ D) is the detailed inspection of the area marked in dashed circle, with A) in the urban area, B) denotes a small station, C) presents a village path, and D) is a nearby park.

*D. Runtime Analysis*

The average runtime for processing each scan in different scenarios is shown in TABLE III, denoting the proposed system capable of real-time operation for all conditions.

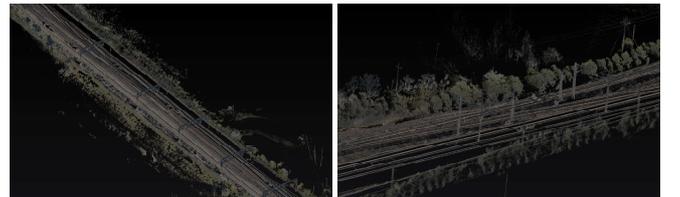

**Fig. 10.** Real-time colored mapping result using the panoramic camera and multiple LiDARs, since the distortion of fisheye camera is large with increased distance, we only take the points within 50 m for mapping.

TABLE III
THE AVERAGE TIME CONSUMPTION IN MS

|  | Preprocessing | LiDAR Frontend | LiDAR Backend | LiDAR Mapping |
|---|---|---|---|---|
| *FY-Back1* | 8.72 | 29.38 | 69.03 | 293.68 |
| *HQ-to2* | 9.23 | 28.75 | 73.23 | 326.85 |

## VI. CONCLUSION

In this paper, we proposed an accurate and robust localization and mapping framework for rail vehicles. Our system integrates measurements from multiple LiDARs, IMU, train odometer, and GNSS in a tightly-coupled manner. Besides, we leverage geometric structure constraints to cope with the rotational divergence due to limited FoV. The proposed method has been extensively validated in large-scale railway, with a decimeter-scale accuracy in most scenarios.

Future work will integrate the panoramic visual information into pose estimation as well as map construction, and verify the system robustness in presence of GNSS failure.

ACKNOWLEDGMENT

We would like to thanks colleagues from Hefei power supply section, China Railway, for their kind support.